\documentclass[]{spie}  

\usepackage{amsmath,amsfonts,amssymb}
\usepackage{graphicx}
\usepackage[colorlinks=true, allcolors=blue]{hyperref}

\title{Synthetic Data-Driven Multi-Architecture Framework for Automated Polyp Segmentation Through Integrated Detection and Mask Generation}

\author[a]{Ojonugwa Oluwafemi Ejiga Peter}
\author[b]{Akingbola Oluwapemiisin}
\author[a]{Amalahu Chetachi}
\author[b]{Adeniran Opeyemi}
\author[b]{Fahmi Khalifa}
\author[a]{Md Mahmudur Rahman}
\affil[a]{Department of Advanced Computing, Morgan State University, Baltimore, Maryland, USA}
\affil[b]{Department of Computer and Electrical Engineering, Morgan State University, Baltimore, Maryland,
USA}

\authorinfo{Further author information: (Send correspondence to O.O.E.P)
\\O.O.E.P: E-mail: ojeji1@morgan.edu\\  
  A.O.: E-mail: olaki62@morgan.edu\\
  A.C.: E-mail: chama6@morgan.edu\\
  A.O.: E-mail: opade7@morgan.edu\\
  F.K : Email: fahmi.khalifa@morgan.edu\\
  M.M.R: E-mail: md.rahman@morgan.edu}

\begin{document} 
\maketitle

\begin{abstract}
Colonoscopy is a vital tool for the early diagnosis of colorectal cancer, which is one of the main causes of cancer-related mortality globally, hence it is deemed an essential technique for the prevention and early detection of colorectal cancer. The research introduces a unique multidirectional architectural framework to automate polyp detection within colonoscopy images while helping resolve limited healthcare dataset sizes and annotation complexities. The research implements a comprehensive system that delivers synthetic data generation through Stable Diffusion enhancements together with detection and segmentation algorithms. This detection approach combines Faster R-CNN for initial object localization while the Segment Anything Model (SAM) refines the segmentation masks. The faster R-CNN detection algorithm achieved a recall of 93.08\% combined with a precision of 88.97\% and an F1 score of 90.98\%.SAM is then used to generate the image mask. The research evaluated five state-of-the-art segmentation models that included U-Net, PSPNet, FPN, LinkNet, and MANet using ResNet34 as a base model. The results demonstrate the superior performance of FPN with the highest scores of PSNR (7.205893) and SSIM (0.492381), while UNet excels in recall (84.85\%) and LinkNet shows balanced performance in IoU (64.20\%) and Dice score (77.53\%). This framework achieves its primary breakthrough through a synthetic data system and an automatic ground truth generator, as these methods combat data limitations without sacrificing medical accuracy. The framework unites multiple architectures together with extensive evaluation metrics to set new benchmarks, which should speed up medical image segmentation tools across different medical specialties.
\end{abstract}


\section{INTRODUCTION}
\label{sec:intro}  
Colonoscopy detects colorectal cancer at an early stage, reducing what is known as one of the main causes of cancer-related deaths and improving patient treatment outcomes ~\cite{bray_2018_global}. A skilled endoscopist makes colonoscopy tests much better in finding polyps, although colonoscopy remains the most important examination for this purpose. Modern deep learning advances show they can better identify polyps and create more precise findings ~\cite{wei_2021_shallow}.The search for automated systems that can accurately detect polyps faces major technical challenges due to data scarcity. Experts must annotate large datasets privately while following patient privacy rules that restrict data access to make meaningful progress. Jha et al. ~\cite{jha_2019_resunet} examined existing datasets that lack polyp samples that reflect the true spectrum of clinical findings because they contain unequal representations of different types of polyps and imaging backgrounds. Research now studies various ways to solve these challenges. Based on Fan et al. ~\cite{fan_2020_pranet} and Zhang et al. ~\cite{qu_2022_transfuse} research showed that semi-supervised learning with unlabeled data can work well, plus transfer learning from natural images shows great results. To address poor results in multiple polyp size and appearance conditions, Liu et al.~\cite{huang_2021_hardnetmseg} built a system that uses multiple levels of image attention.

This research offers a novel approach by presenting a synthetic data-driven framework that makes use of the most recent developments in AI-generated imaging. Using DreamBooth low-rank Adaptation (LoRA) and fine-tuned Stable Diffusion, we create a varied artificial collection of colonoscopy images. The framework has a multistage pipeline that uses Five Models for Image Segmentation, the Segment Anything Model (SAM) for accurate mask generation, and Faster R-CNN for early detection.
The proposed framework includes five cutting-edge architectures: U-Net, PSPNet, FPN, LinkNet, and MANet, each with a ResNet34 backbone pre-trained in ImageNet. To ensure a robust performance assessment, these models are tested using a wide range of measures, including IoU, Dice coefficient, and F1.
The significance of this research lies in its potential to:
\begin{enumerate}
    \item Innovative methods are needed to solve the medical image segmentation issue of limited patient datasets
    
    \item Develop a usable method to make accurate training data from scratch effectively
     \item To evaluate and compare the performance of five state-of-the-art segmentation models on this synthetic 
        dataset. 
    \item Assess the potential of synthetic data in improving the accuracy and reliability of colonoscopy images
segmentation
\end{enumerate}

\section{RELATED WORK}
Medical Image Segmentation stands as a vital healthcare tool that partitions medical images to help doctors locate structural details and disease patterns during diagnosis and treatment preparation. The medical image segmentation field has developed rapidly from hand labeling to advanced computerized systems. Prasantha et al.~\cite{s_2010_medical} explain how basic thresholding and edge detection laid the groundwork for automated segmentation, but also showed its first fundamental principles. The integration of machine learning further developed our medical practices. Ahlwar and Jadon ~\cite{ahirwar_2015_fcsofm}  implemented FCSOFM in 2015, allowing better detection of breast abnormalities by MRI. During this time, researchers achieved excellent results when they used clustering techniques and Support Vector Machines to spot tumors and type body tissues. New deep learning approaches have completely changed the way medical images are segmented. According to Daza et al. ~\cite{daza_2021_towards} research published in 202,1 their work demonstrated how U-Net with DenseNet blocks and ROG lattice architecture with robust adversarial training enhanced medical brain tumor segmentation. New methods have successfully enhanced segmentation results across the fields of medicine, including cancer treatment and heart and brain imaging. Although progress has been made, the fields of medical image analysis still face complex issues. Limited access to quality datasets remains the main challenge for medical image analysis as institutions follow different diagnostic protocols. Das and Das ~\cite{das_2020_medical}  showed that high-resolution medical image processing requires large amounts of computer power to function well during real clinical workflows. Researchers encounter difficulties in building defense systems that block tampered images, and they also work well for multiple medical conditions.

Jha et al. ~\cite{jha_2019_resunet} created ResUNet++ by combining residual learning and squeeze-and-excitation blocks with ASPP in the UNet architecture. By evaluating their method on Kvasir-SEG and CVC-ClinicDB data, they produced accuracy scores of 0.813 and 0.796. Their work showed promise in learning new features, yet required better performance in both the instant and when detecting small polyps. Fan et al.~\cite{fan_2020_pranet} created PraNet to improve boundary analysis through parallel reverse attention. In the Kvasir-SEG medical images, they produced a 95 FPS connection with a 0.898 accuracy rating. The method performed poorly when faced with polyps of different sizes in challenging background scenes. Borgli et al.~\cite{borgli_2020_hyperkvasir} launched HyperKvasir, which holds more than 110,000 endoscopic pictures. This dataset helped researchers make progress while showing problems with unreliable labeling and unequal data distribution across categories. We want our synthetic data collection method to help existing real datasets by providing reliable tagging information. Zhou et al.~\cite{zhou_2019_unet} designed UNet++ to enhance feature preservation by densely connecting information layers. Their network produced an accuracy of 0.818 while necessitating significant computing power. They demonstrated that merging features correctly leads to better performance.

Liu et al.~\cite{huang_2021_hardnetmseg} combined an efficient neural network with multiscale processing to create HarDNet-MSEG. Their work reached 45 FPS processing speed with a 0.912 Dice score in CVC-ClinicDB, demonstrating the best performance and speed ratio yet. Zhang et al.~\cite{qu_2022_transfuse} created TransFuse to become one of the first successful applications to use transformers in polyp segmentation. Their hybrid CNN-transformer model generated 0.902 Dice scores on Kvasir-SEG and showed that attention methods work well in medical image segmentation tasks. Despite using more computational power, the model showed poor accuracy results in all testing datasets.

 Jha et al.~\cite{jha_2021_kvasirinstrument} developed the Kvasir instrument, which segmented both the polyp and the instrument in endoscopic images. They provided a baseline for the segmentation of the GI tools to promote research and algorithm development. They obtained a dice coefficient score of 0.9158 and a Jaccard index of 0.8578 using a classical U-Net architecture. Tomar et al.~\cite{tomar_2020_ddanet} introduced DDNet to help better detect boundaries with two separate decoding systems. Their network produced a 0.894 Dice score on CVC-ClinicDB while showing better detection for polyp boundaries, but needed independent training stages to achieve its best outcome.

Our extensive research shows that current studies have fundamental gaps in the field. Researchers have not devoted enough attention to using synthetic data to improve learning, while medical imaging remains difficult to obtain. Researchers lack consistent ways to test different methods across many databases, which hinders their ability to compare results. Our research shows the need for better medical tools that handle the resource limitations found in healthcare facilities. The research community does not fully explore domain adaptation methods as a solution to combine synthetic and real medical imaging datasets. Our study presents multiple key enhancements to medical image processing technology. This research creates a new method to create high-quality synthetic medical image data for better training of analysis systems. This research introduces a complete evaluation system that enables defined standard testing for multiple deep learning methods. Our method designs specific network architectures and optimization methods to provide a realistic combination of speed and accuracy that works best in practical medicine. Our system combines detection and segmentation functions into one integrated platform to simplify analysis tasks while preserving excellent performance results.

\section{RESEARCH METHODOLOGY}

\begin{figure}[ht]
   \begin{center}
   \begin{tabular}{c}
   \includegraphics[width=0.95\textwidth, height=20cm, keepaspectratio]{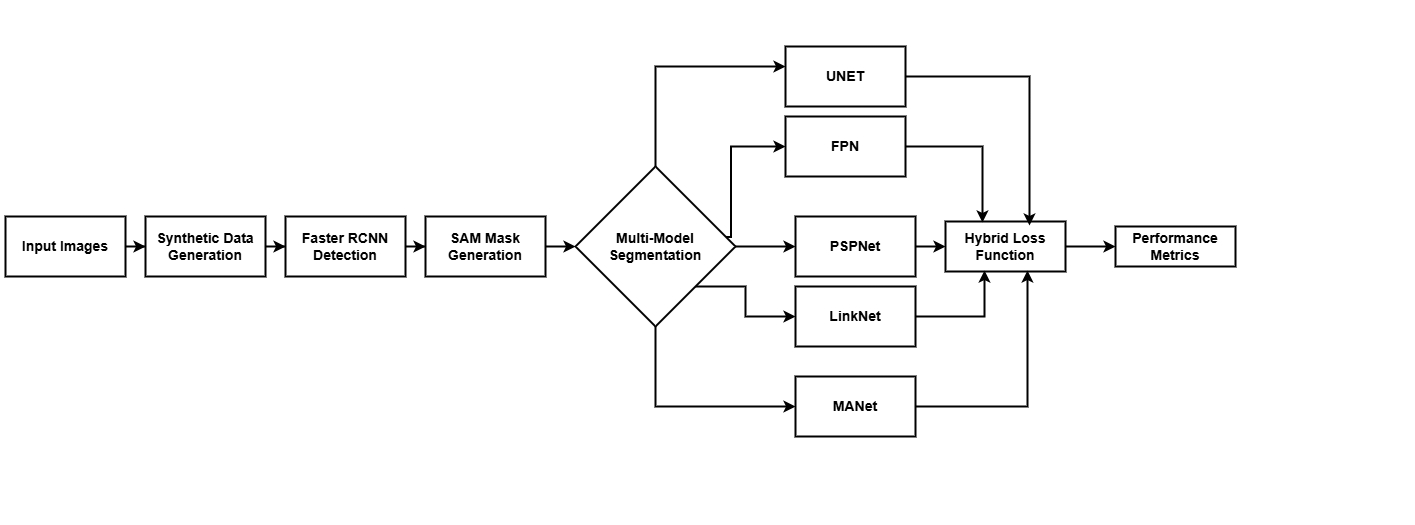}
   \end{tabular}
   \end{center}
   \caption[example2]
   { \label{fig:fig1} 
Pipeline diagram showing medical image processing workflow: from input through synthetic data generation, R-CNN detection, SAM mask generation, multi-modal segmentation (using U-Net, PSPNet, FPN, LinkNet, and MANet), hybrid loss function, and performance metrics.}
\end{figure}

The method explains how to create an entire automated polyp segmentation system that combines synthetic image generation with powerful deep learning models. This method creates a series of steps to develop synthetic colonoscopy images by using Stable Diffusion and DreamBooth LoRA to adjust and generate new images. By creating synthetic medical images, our system solves dataset insufficiency and provides both varied training data and accurate labels. The system performs two-stage image analysis starting with Faster R-CNN ~\cite{ren_2015_faster} to find specific areas of interest that SAM~\cite{kirillov_2023_segment} uses to generate accurate segmentation results. Our setup checks U-Net~\cite{ronneberger_2015_unet}, PSPNet~\cite{zhao_2017_pyramid}, FPN~\cite{lin_2017_feature}, LinkNet~\cite{chaurasia_2017_linknet}, and MANet~\cite{fan_2020_manet} segmentation methods that work with ResNet34 and have been pre-trained on ImageNet. The research combined BCE, Dice, and Focal losses for an effective result analysis. The system trains on extensive data modification with an 80:20 split between train and test while providing detailed performance measurements that include IoU, F1 score, Dice coefficient, PSNR and SSIM values. Our complete method advances medical image segmentation and gives doctors a reliable approach to detecting and segmenting polyps for use in the real world.  
 Medical image segmentation requires robust training pipelines and evaluation frameworks. The diagram in Fig.~\ref{fig:fig1} illustrates the complete workflow from data preparation through model training to final evaluation.
\newpage

\subsection{Data Flow Pipeline}
\label{sec:title}
The system starts with the Data Flow Pipeline, which turns several processed steps into training data for segmentation tasks. It processes synthetic images as input data and runs them through a faster R-CNN detection tool. Our system detection phase finds areas of interest (potential polyp spots) in all images directly. The Faster R-CNN model, built on a ResNet50 backbone, generates two types of outputs: Distance values for boxes are stored as text data while face locator models generate image results by adding visible box-outlines. The detection results act as a link between the raw images and the segmented areas so that the system can work its way toward producing a final result. The Faster R-CNN classifier is trained using the CVB dataset, and the inference is done using the synthetic data generated using Fine-tuned Stable diffusion and DreamBooth LoRa to produce two classes of synthetic images, namely Polyps and Non-Polyps Class.

\begin{equation}
\label{eq:detection}
D_{RCNN}(I) = \sum_{i=1}^{n} \{B_i, P_i\} = \{(x_i, y_i, w_i, h_i), conf_i\}
\end{equation}

where $D_{RCNN}(I)$ processes the input image $I$, generating bound boxes $B_i$ with coordinates $(x_i, y_i, w_i, h_i)$ and confidence scores $conf_i$ for the detected regions $n$.

After finding objects, the pipeline uses SAM to process the paired images. The Segment Anything Model processes paired images and text annotations. The specific text boxes tell SAM exactly where to draw its accurate segmentation lines. The system turns each box prediction into detailed polyp masks to show exactly where the polyps live on the screen. SAM functions as both a connection and an enhancement point between detection and segmentation systems through its prompt-based segmentation methods to develop precise ground-truth masks.
\begin{equation}
\label{eq:sam_mask}
M_{SAM}(x,y) = \sigma(\phi(f(I), p(B)) \cdot E(x,y))
\end{equation}
where $\varphi$ represents SAM's embedding function, f(I) is image features, p(B) is prompt from bounding boxes, and E(x,y) is positional encoding.

The data organization structure keeps everything in Google Drive in a properly arranged state. The directory setup keeps synthetic images (/synthetic) separate from two more directories: /masks for generated outlines and /SAM-Results for SAM output. Alongside these are the results and synthetic images folders (/results and /synthetic). The systematic storage scheme lets personnel handle data properly between pipeline phases and simplifies access to all types of information.

\subsection{Model Architectures}
\label{sec:title}

The system employs five separate segmentation models, which we display through the Model Architectures diagram in Fig.~\ref{fig:fig1}. Each model has specific architecture components and certain shared system functions. All of these models share common components: The model uses a ResNet34 architecture trained on ImageNet for strong feature extraction, plus three loss functions: Binary Cross Entropy, Dice, and Focal, while training with AdamW Optimizer, which includes automatic weight decay. Equation (3) shows the Unified Framework for Polyp Segmentation:

\begin{equation}
\label{eq:unified_architecture}
\Phi(I) = \underbrace{\mathcal{M}_k}_{\text{segmentation}} \Big( \underbrace{\mathcal{S}}_{\text{SAM}} \big( \underbrace{\mathcal{D}(I)}_{\text{detection}} \big) \Big), \quad k \in \{U, P, F, L, MA\}, \quad \min_{\theta} \{\lambda_1\mathcal{L}_{det} + \lambda_2\mathcal{L}_{seg} + \lambda_3\|\theta\|_2\}
\end{equation}

The unified equation $\Phi(I)$ encapsulates our complete polyp segmentation pipeline, where an input image $I$ undergoes three sequential stages:

\begin{enumerate}
   \item The detection function $\mathcal{D}(I)$ utilizing Faster R-CNN identifies potential polyp regions
   \item The SAM processing function $\mathcal{S}$ refines the detected regions into precise masks
   \item The segmentation function $\mathcal{M}_k$ performs final segmentation using one of five architectures: $k \in \{U, P, F, L, MA\}$ representing U-Net, PSPNet, FPN, LinkNet, and MANet, respectively
\end{enumerate}

The system is optimized through the minimization of a combined loss function:
\begin{enumerate}
   \item $\mathcal{L}_{det}$: Detection loss ensuring accurate polyp localization
   \item $\mathcal{L}_{seg}$: Segmentation loss guaranteeing precise mask generation
   \item $\|\theta\|_2$: L2 regularization term preventing overfitting
   \item $\lambda_1, \lambda_2, \lambda_3$: Weighting coefficients that balance the contribution of each loss component
\end{enumerate}

The segmentation framework combines five architectural strategies to work best for polyp segmentation. The U-Net system functions through its core encoder-decoder design and skip connections that retain spatial precision at all image resolutions. Through its pyramid-pooling design, PSPNet delivers context information from multiple spatial resolutions for better feature comprehension. FPN leads multi-scale processing by designing a feature pyramid that synchronizes fine image details with rich semantic content across different levels of connections. LinkNet delivers computing speed improvements by merging a simple decoder setup with direct residual blocks, yet keeps segmentation results stable. MANet improves the system by helping the model discover critical polyp features in a process that eliminates background noise from attention blocks. These segmentation approaches work together by tackling different aspects of polyp detection, including spatial representation and performance optimization.

\subsection{Training and Inference Pipeline}
\label{sec:title}

\begin{figure}[ht]
   \begin{center}
   \begin{tabular}{c} 
   \includegraphics[height=5cm]{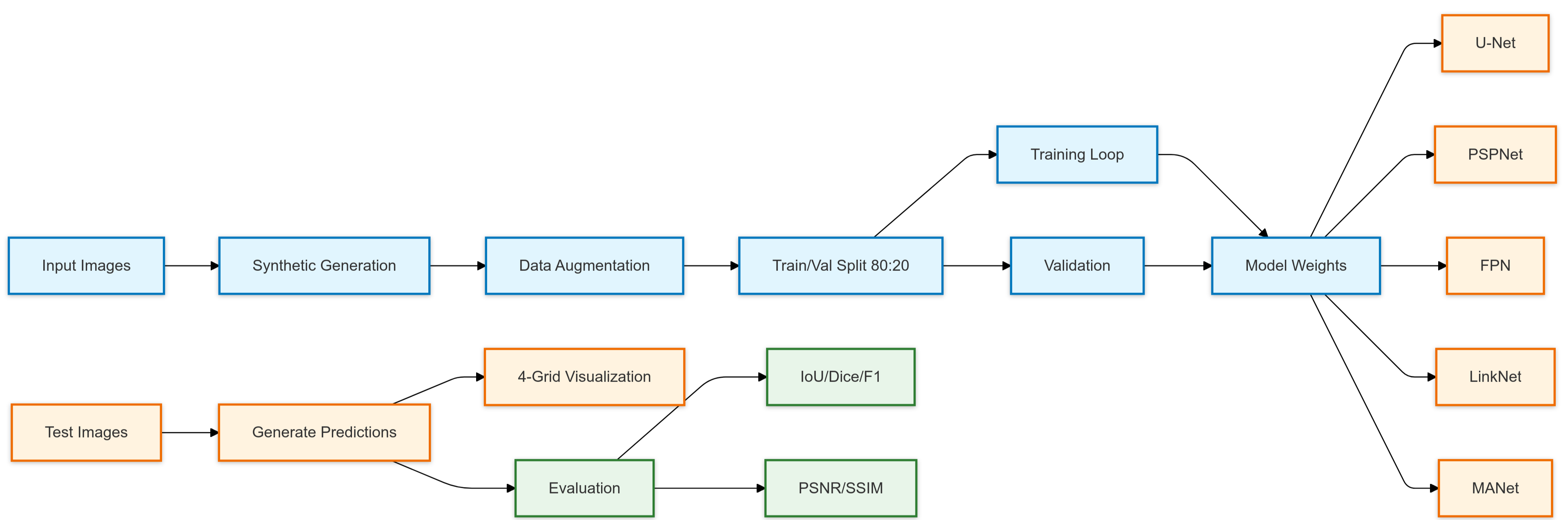}
   \end{tabular}
   \end{center}
   \caption[example2]{\label{fig:fig2} Training and evaluation pipeline for medical image segmentation: Input processing through synthetic data generation and augmentation, model training with validation (80:20 split), and comprehensive evaluation using multiple architectures (U-Net, PSPNet, FPN, LinkNet, MANet) with quality metrics (IoU/DiceF1, PSNR/SSIM) and 4-Grid visualization.}
\end{figure}

The training procedure uses a dataset that receives multiple enhancements through random rotations, flips, distorted size modifications, and brightness changes. The expanded data set receives an 80:20 split between the training and validation sets for model evaluation during training. The training loop consists of four key steps: the training pipeline performs a forward pass, computes the hybrid loss, calculates gradients in reverse mode, and then updates weights using the chosen optimizer. When training stops, the model validation results select the model with the best performance, so we save those weights. Training proceeds through several loops with epochs as each cycle checks performance on the validation set to stop growth when needed. A trained model performs polyp classification tasks using new images through the inference pipeline. Our system starts by loading the model weights and then passes test images through every stage of the model. The prediction process generates segmentation masks that are then visualized in a standardized 1x4 grid format: The output system displays both the input image and separate views that show the model's segmentation findings and outcomes compared to ground truth. The design structure covers all necessary steps for processing data, from preparation to training and performance assessment. The system's module-based design enables quick maintenance and update work, while its performance evaluation system shows us what our model does. The workflow structure helps researchers create repeatable results and shows how the segmentation steps work, which is beneficial for researchers and clinicians working with polyp images.Fig.~\ref{fig:fig2} shows that the segmentation of medical images requires robust training pipelines and evaluation frameworks. The diagram illustrates the complete workflow from data preparation through model training to final evaluation. 

\section{RESULT}

\begin{figure} [ht]
   \begin{center}
   \begin{tabular}{c} 
   \includegraphics[height=5cm]{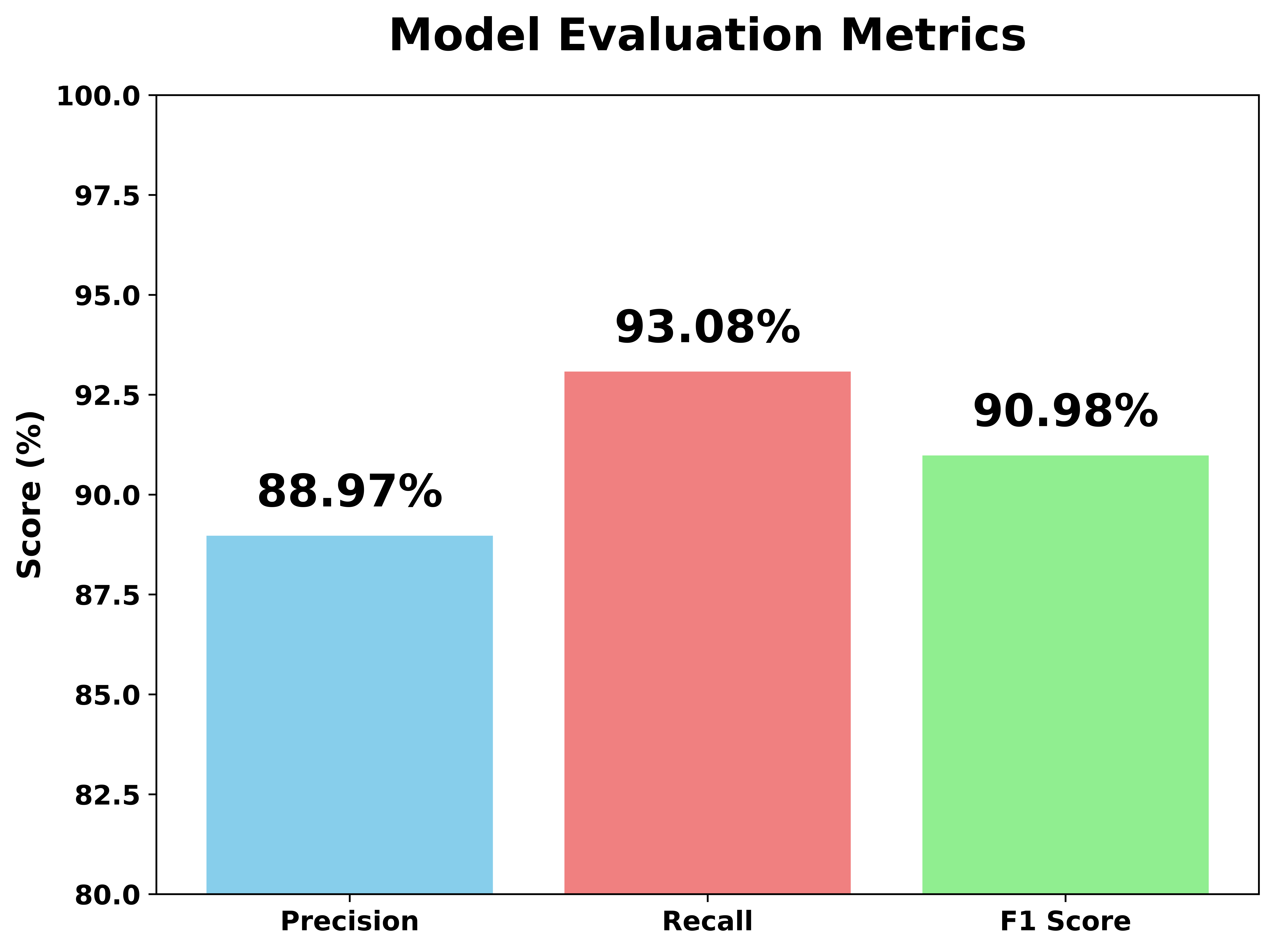}
   \end{tabular}
   \end{center}
   \caption[example3] 
   { \label{fig:fig3} 
Model Evaluation Metrics showing performance across three key metrics: Precision (88.97\%), Recall (93.08\%), and F1 Score (90.98\%), visualized as a bar chart with blue, red, and green bars respectively.}
   \end{figure}

\begin{figure}[ht]
   \begin{center}
   \begin{tabular}{c}
   \rotatebox{90}{\includegraphics[height=15cm]{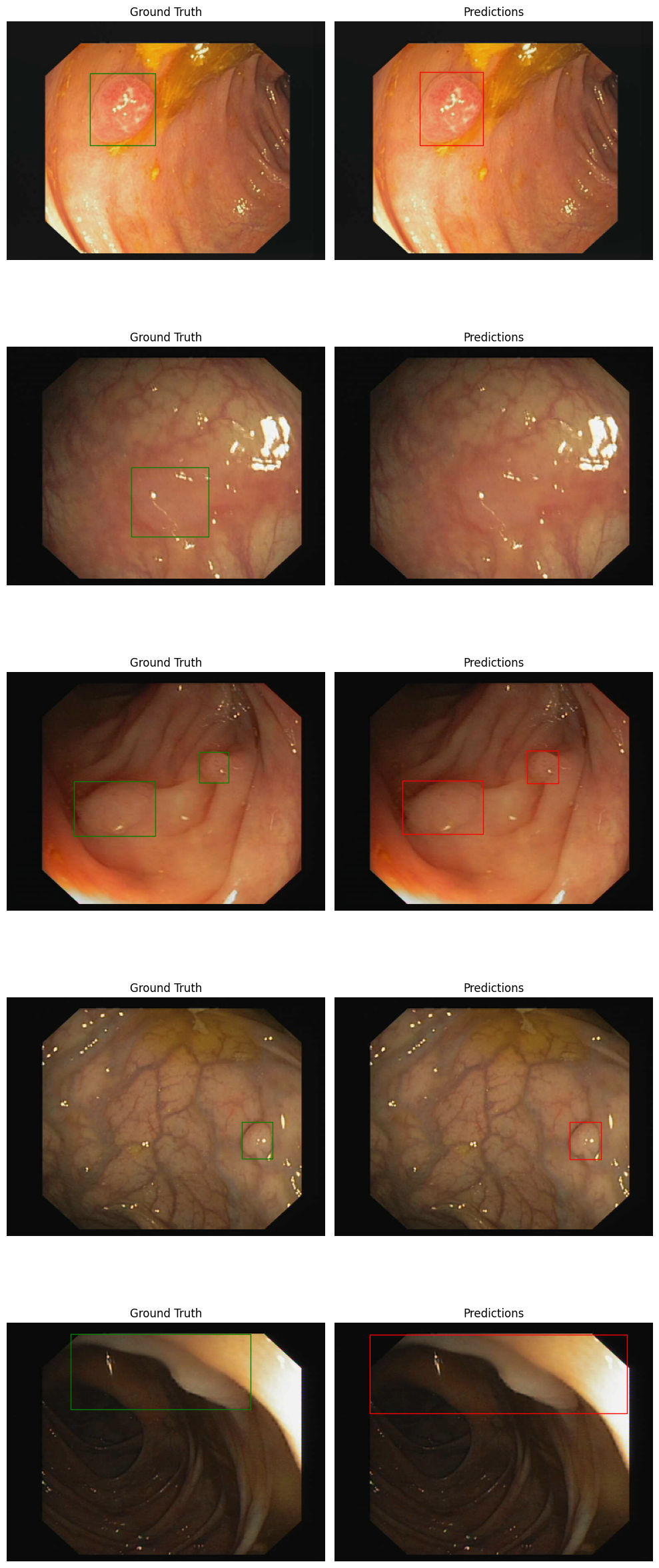}}
   \end{tabular}
   \end{center}
   \caption[example4]
   { \label{fig:fig4} 
Comparison of Faster R-CNN polyp detection results: Original endoscopic images (top row) and detection outputs (bottom row) showing accurate polyp localization (red bounding boxes) across different lighting conditions, angles, and tissue appearances.}
\end{figure}

A faster analysis of the R-CNN framework reveals detection metrics that confirm a precision level of 88.97\% for correct polyp detection with low false positive errors. For medical screening applications, the recall rate is 93.08\% ensures that the system effectively finds the most genuine polyps in the images while reducing the risks of dangerous false negative results. The F1 score at 90.98\% expresses the optimal management of the accurate prediction model and total identification using a precision, recall harmonic mean calculation.
 Fig.~\ref{fig:fig3} shows the performance metrics of the faster R-CNN polyp detection model with precision (88.97\%), recall (93.08\%), and F1 score (90.98\%), showing strong detection capabilities with a balanced precision-recall trade-off.
Multiple performance metrics track detection phase capabilities within a multi-architecture automated framework through Faster R-CNN operating with ResNet50 backbone topology. The initial step screens for potential polyps and provides corner points that guide subsequent processes for both detailed boundary identification and refined segmentation. The high accumulated values across all metrics confirm that this tool functions properly as an initial assessment method despite having undergone precise adjustments meant to better suit medical image analysis requirements that require optimal detection levels while avoiding unnecessary false alarm signals.
 Fig.~\ref{fig:fig4} Faster R-CNN demonstrates effective polyp detection under various endoscopic conditions, as shown by these validation results.

\begin{figure}[ht]
   \begin{center}
   \begin{tabular}{c}
   \includegraphics[height=7cm]{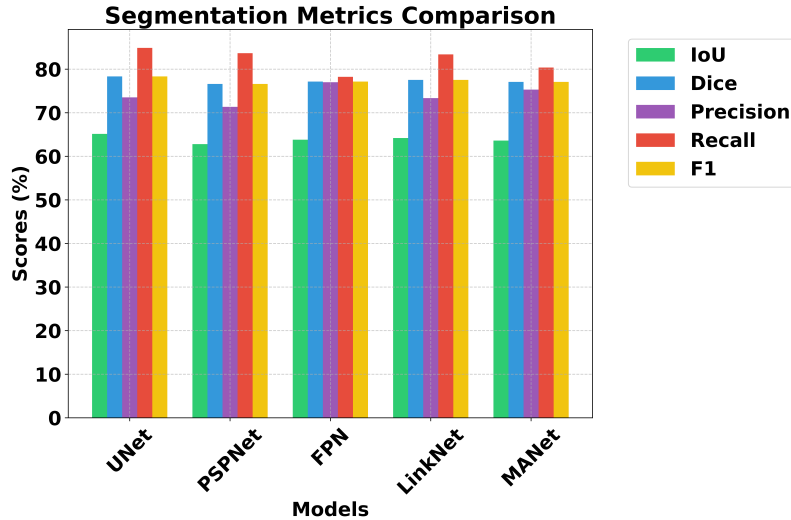}
   \end{tabular}
   \end{center}
   \caption[example5]
   { \label{fig:fig5} 
   Performance comparison of segmentation models (U-Net, PSPNet, FPN, LinkNet, MANet) evaluated using multiple metrics: IoU, Dice coefficient, Precision, Recall, and F1 score. All models achieved above 60\% IoU and more than 70\% across other metrics, with U-Net showing marginally better performance.}
\end{figure}

FPN (Feature Pyramid Network) stands as the best solution for polyp segmentation operations based on analysis across six evaluation metrics. The FPN architecture generates optimal image quality by reaching the highest PSNR value of 7.205893 and the SSIM score of 0.492381 for segmenting medical images. The precision results for FPN are 77.00\%, indicating a robust boundary detection accuracy for polyps.  Fig.~\ref{fig:fig5} shows the valuation metrics and indicates consistent performance across architectures, with slight variations in precision-recall trade-offs.

\begin{figure}[ht]
   \begin{center}
   \begin{tabular}{c}
   \includegraphics[height=7cm]{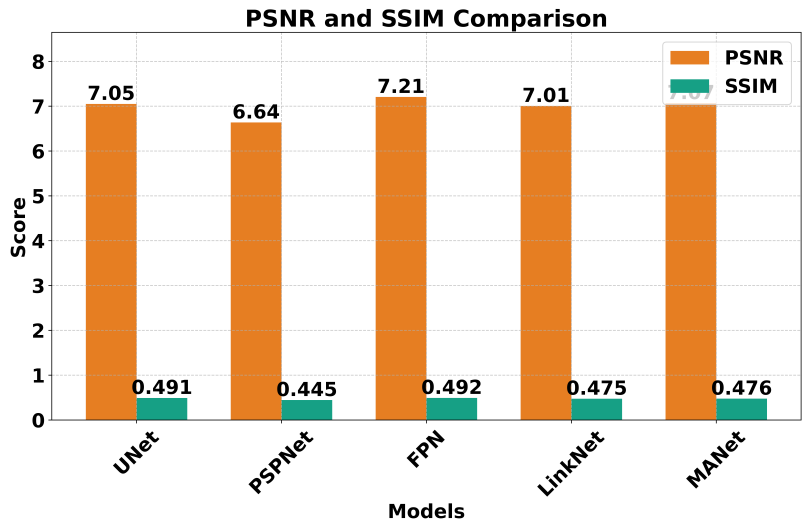}
   \end{tabular}
   \end{center}
   \caption[example5]
   { \label{fig:fig6} 
Higher PSNR values indicate better signal quality relative to noise in the segmented outputs.SSIM values closer to 1 indicate better preservation of structural information in segmentation outputs.}
\end{figure}

The UNet model exhibits solid recall performance (84.85\%) in addition to maintaining consistent metric scores. FPN exhibits outstanding recall efficiency because it detects all potential polyp areas at the cost of somewhat reduced precision compared to network FPN. The model's PSNR result (7.050282) is equivalent to other models while keeping its performance metrics evenly balanced.  Fig.~\ref{fig:fig6} shows the comparison of PSNRs between segmentation models, with FPN achieving the highest score (7.21), followed by MANet (7.07) and U-Net (7.05). PSPNet showed the lowest PSNR at 6.64.
LinkNet achieves evenly distributed results across all metrics while achieving its strongest outcomes through the IoU (64.20\%) and the Dice score (77.53\%). The algorithm's consistent metrics signal its ability to generalize and consistently segment a variety of polyp images properly.  Fig.~\ref{fig:fig6} shows the SSIM measurement between models, with FPN showing the highest structural similarity (0.492), closely followed by U-Net (0.491). PSPNet demonstrated the lowest SSIM score (0.445), while LinkNet and MANet achieved moderate scores of 0.475 and 0.476, respectively.

\begin{figure}[ht]
   \begin{center}
   \begin{tabular}{c}
   \includegraphics[height=5cm]{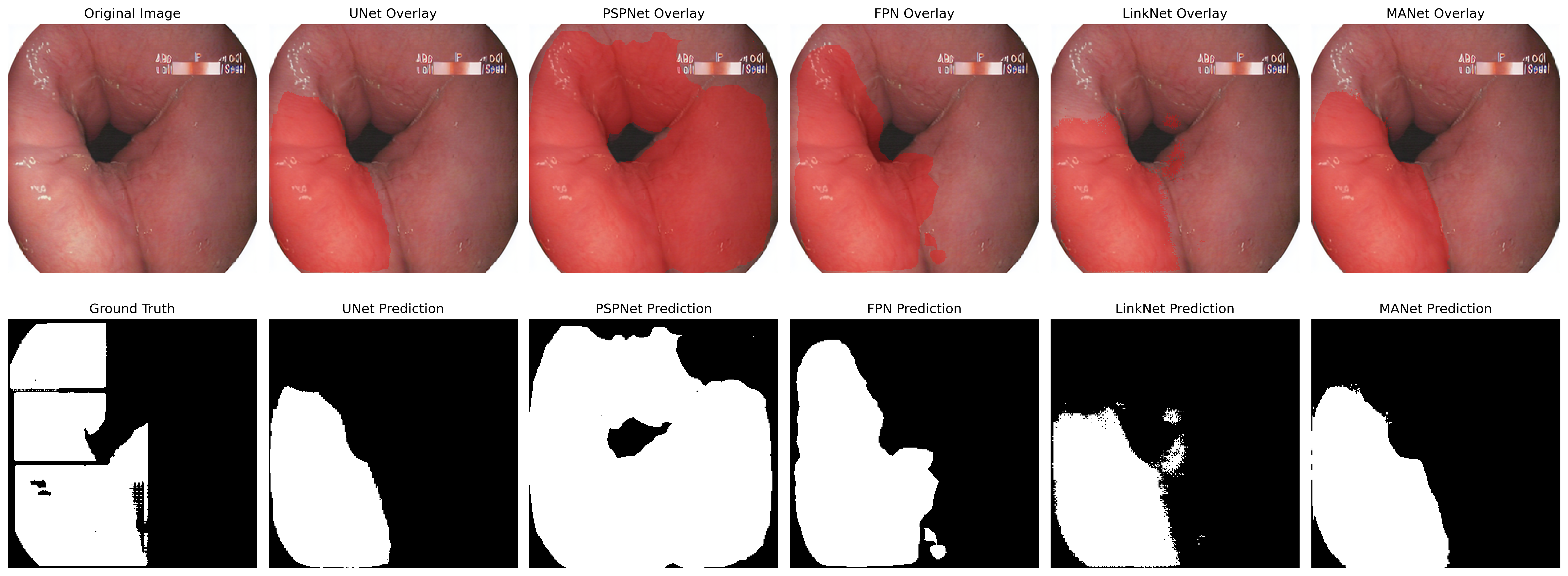}
   \end{tabular}
   \end{center}
   \caption[example4]
   { \label{fig:fig9} 
Comparison of medical image segmentation models: Original endoscopic image with ground truth segmentation mask and predictions from five different deep learning architectures (UNet, PSPNet, FPN, LinkNet, and MANet), showing both overlay visualizations (top) and binary prediction masks (bottom)}
\end{figure}

Although MANet exhibits regular performance on each assessment metric, it does not surpass other methods in any measure. The integration of numbers from different metrics shows steady results (63.81\% for IoU, but also 77.07\% for Dice and other metrics), indicating reliable but modest performance. In this precise application, the attention mechanism integrated into MANet does not demonstrate meaningful benefits. Fig.~\ref{fig:fig9} demonstrates segmentation performance across different architectures.
Analysis shows that PSPNet delivers the worst overall results from all five models due to its underperforming PSNR measurement (6.637351) and SSIM outcome (0.445212). The model maintains an 83.66 percent recall score, which demonstrates it still recognizes polyp regions yet its precision falls below other tested models. FPN proves to be the most recommended model for medical image segmentation, considering both the entire set of evaluation metrics together with essential performance equality requirements. Many clinical applications value FPN because it delivers superior image quality performance while maintaining strong precision scores, which ensures crucial boundary detection accuracy. The model architecture demonstrates excellent feature integration capability through its pyramid structure, which provides efficient processing of medical images showing different polyp dimensions. The higher recall ability of UNet makes it attractive for medical applications where the cost of missed polyps outweighs false positives. Whether FPN or UNet gets selected as the medical imaging model depends on how significant precise results matter relative to capturing every important feature within their specific application.

\section{Discussion}
This research features a novel multiple-stage processing pipeline that integrates advanced object detection and universal segmentation models with medical image segmentation technologies. The unified system that joins Faster R-CNN with the Segment Anywhere Model (SAM) develops an automatic ground truth production framework that tackles the major obstacle of scarce medical image labels within clinical research. The resulting hybrid system starts by identifying candidate regions with Faster R-CNN before SAM runs a refinement process for precise segmentation mask generation. The new automated ground truth creation system speeds up annotation tasks without compromising quality standards. The integration of SAM with domain-specific detection capabilities introduces an innovative alliance of general-purpose and specialized medical image processing methods. The research used a unified training and inference system to evaluate five segmentation structures (U-Net, PSPNet, FPN, LinkNet, and MANet) simultaneously through a comprehensive evaluation framework. The use of a developed hybrid loss function that unites BCE with Dice and Focal losses through optimized weights shows promise as a solution for tackling polyp segmentation requirements regarding class frequency discrepancies and border accuracy. The adaptive training strategy in this architecture integrates dynamic learning rate adjustments with mixed precision training approaches. The visualization pipeline introduces a standardized 1x4 grid layout to produce a new method to evaluate the performance of the model among different network architectures. The combined metric detection assessment system utilizes segmentation-specific evaluations (IoU and Dice) alongside image quality metrics (SSIM and PSNR) to offer a more extensive method of performance assessment. This research brings significant value to medical image analysis by offering a fully documented, comprehensive pipeline that improves segmentation performance and addresses crucial operational hurdles with respect to diminished medical images and annotation difficulties.
This research introduces four major innovations in medical image processing that collectively advance automated analysis capabilities. The first breakthrough is a synthetic data generation pipeline that addresses the persistent challenge of limited medical datasets. Built using Stable Diffusion and SAM, this system produces realistic medical imagery while maintaining clinical relevance. Particularly noteworthy is its application in generating synthetic colonoscopy data, which helps alleviate the current shortage of medical training data.
The second innovation involves automated ground-truth generation through a sophisticated combination of Faster R-CNN and SAM within a multi-architecture framework. This hybrid approach represents a significant improvement over traditional manual annotation methods by integrating basic computer vision segmentation with specialized medical detection technology. The system automatically generates precise segmentation masks from detected regions, substantially improving the efficiency of medical image processing. A comprehensive evaluation framework marks the third major advancement. This standardized system incorporates multiple metrics, including IoU, Dice, PSNR, and SSIM to provide detailed insights into model performance. The framework features a dual evaluation mechanism that simultaneously assesses both detection and segmentation capabilities, along with a standardized visualization pipeline. This combination establishes new benchmarks for evaluating segmentation models in medical imaging applications. The fourth innovation centers on multi-model architecture integration, demonstrating the benefits of combining diverse approaches to medical image analysis. By incorporating five different segmentation architectures within a unified framework, the research leverages the unique strengths of each model to comprehensively address polyp segmentation challenges. The framework's modular design ensures reproducibility while facilitating quick updates and maintenance.
These innovations collectively transform the field of medical image analysis in several crucial ways. They effectively address the database limitations through synthetic data generation, enable more efficient annotation processes through automated ground truth generation, establish new evaluation standards and benchmarks, and demonstrate the power of integrating multiple architectural approaches for optimal results. These research advances create opportunities for the application of medical imaging technology in different diagnostic procedures beyond their initial applications in polyp segmentation. The synthetic data created by this framework shows promise for training medical imaging systems in all clinical domains as a result of its ability to maintain realistic patient data consistency.
The study enables three main directions for future advancement, which comprise time-sensitive polyp detection methods, higher medical diagnostic precision, and reduced requirements for physician annotation of medical images. The proven effectiveness of synthetic data to train segmentation models indicates new opportunities for addressing data shortages in various medical imaging fields, which may expedite the development of automatic diagnostic tools throughout all medical specialties.
The research does not address real-time implementation capabilities, which are crucial for clinical deployment. The attention mechanism in MANet showed low performance, indicating limitations in its effectiveness for this specific medical context. The study also lacks exploration of domain adaptation methods for synthetic-to-real data transfer. Additionally, the research needs further investigation into knowledge transfer between different medical imaging datasets and automated methods to enhance model performance across various clinical applications

\section{Conclusions and Future Work}
This research develops an advanced medical image segmentation system by combining synthetic data generators with automated polyp detection and segmentation technology inside a multi-architecture framework. Medical imaging technology reached a breakthrough because the system combines Stable Diffusion and Segment Anything Model (SAM) with five state-of-the-art segmentation models to generate amazing results in colonoscopy image analysis.
The faster R-CNN provided exceptional performance during the detection stage with recall metrics of 93.08\% and precision of 88.97\% along with a 90.98\% F1 score to establish robust initial polyp detection capabilities. A high-performance detection system provides essential support in medical screening procedures that prioritize the complete elimination of incorrect negatives to protect patient safety.
The tests revealed that FPN emerged as the top among segmentation systems due to its consistent performance in all assessment metrics. The combination of PSNR (7.205893) and SSIM (0.492381) results together with the precision of boundary detection (77.00\%) demonstrates the effectiveness of FPN for medical image segmentation. The pyramid-shaped design of this architecture functions best with the various dimensional and form characteristics that appear in medical image segmentation applications.
The UNet model exhibits solid recall effectiveness at 84.85\% along with stable metric results that established it as an available selection, specifically when complete polyp region identification rates priority over precision accuracy. LinkNet maintains competitive performance between its IoU score of 64.20\% and the Dice score of 77.53\%, demonstrating that it can handle a wide range of different polyp images successfully. However, MANet exhibits consistent results but mediocre performance in all metrics, indicating that the attention mechanism does not provide noticeable advantages in this specific context.
Our extensive evaluation framework, which includes several performance metrics, gives physicians unmatched knowledge about the functioning and effectiveness of the model. The combination of segmentation-specific measurements (IoU and Dice) and image quality criteria (SSIM and PSNR) introduces an advanced method to evaluate the performance standards of medical image analysis. Future work needs to focus on both real-time system development and model speed enhancements for deployment in clinical settings and domain adaptation methods to enhance synthetic-real data cross-transference. Research needs to investigate how models transfer knowledge between different medical imaging data sets while simultaneously developing automated methods to enhance model performance throughout different clinical applications.

\acknowledgments 
 
This work was supported by the National Science Foundation (NSF) grant (ID. 2131307) “CISE-MSI: DP:IIS: III: Deep Learning-Based Automated Concept and Caption Generation of Medical Images to Develop Effective Decision Support". The data presented in this study are openly available. The source code and implementation details can be found in \href{https://github.com/Ejigsonpeter/Text-Guided-Synthesis-for-Colon-Cancer-Screening/tree/main}{GitHub Repository}. The datasets used in this study are accessible through secure cloud storage. The complete dataset collection is available in \href{https://drive.google.com/drive/folders/1MJVT0yRlz0mfwbxLWTqCD7idRcTV3hIY?usp=sharing}{Dataset Repository}

\bibliographystyle{spiebib} 

\end{document}